\documentclass[11pt,a4paper,fleqn]{article}
\usepackage[hyperref]{emnlp-ijcnlp-2019}
\usepackage{times}
\usepackage{latexsym}
\usepackage{booktabs}
\usepackage{url}
\usepackage{multicol}
\usepackage{amsmath}
\usepackage{float}
\usepackage[pdftex]{graphicx}
\usepackage{multirow}
\usepackage{subcaption}
\usepackage{algorithm}
\usepackage[noend]{algpseudocode}

\aclfinalcopy 

\title{Adversarial Learning with Contextual Embeddings for Zero-resource Cross-lingual Classification and NER}
\author{Phillip Keung, Yichao Lu, Vikas Bhardwaj \\Amazon Inc. \\{\tt \{keung, yichaolu, vikab\}@amazon.com}\\}

\date{}

\begin{document}
\maketitle
\begin{abstract}
	Contextual word embeddings (e.g. GPT, BERT, ELMo, etc.) have demonstrated state-of-the-art performance on various NLP tasks. Recent work with the multilingual version of BERT has shown that the model performs very well in zero-shot and zero-resource cross-lingual settings, where only labeled English data is used to finetune the model. We improve upon multilingual BERT's zero-resource cross-lingual performance via adversarial learning. We report the magnitude of the improvement on the multilingual MLDoc text classification and CoNLL 2002/2003 named entity recognition tasks. Furthermore, we show that language-adversarial training encourages BERT to align the embeddings of English documents and their translations, which may be the cause of the observed performance gains.
\end{abstract}

\section{Introduction}

Contextual word embeddings \cite{bert, elmo, gpt} have been successfully applied to various NLP tasks, including named entity recognition, document classification, and textual entailment. The multilingual version of BERT (which is trained on Wikipedia articles from 100 languages and equipped with a 110,000 shared wordpiece vocabulary) has also demonstrated the ability to perform `zero-shot' or `zero-resource' cross-lingual classification on the XNLI dataset \cite{xnli}. Specifically, when multilingual BERT is finetuned for XNLI with English data alone, the model also gains the ability to handle the same task in other languages. We believe that this zero-resource transfer learning can be extended to other multilingual datasets.

\begin{figure*}[ht]
	\centering
	\begin{subfigure}[t]{0.6\textwidth}
		\centering
		\includegraphics[width=11cm]{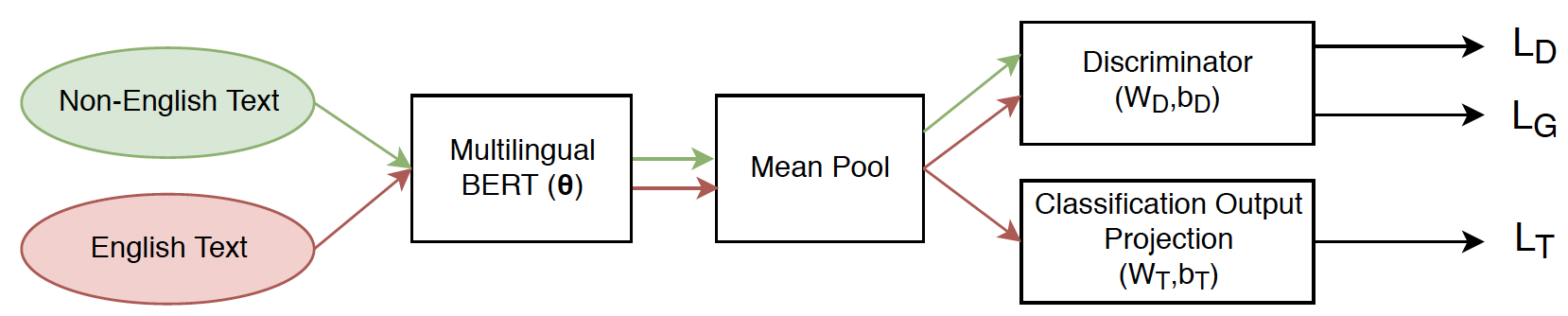}
		\caption{Text classification}
	\end{subfigure}
	\begin{subfigure}[t]{0.6\textwidth}
		\centering
		\includegraphics[width=11cm]{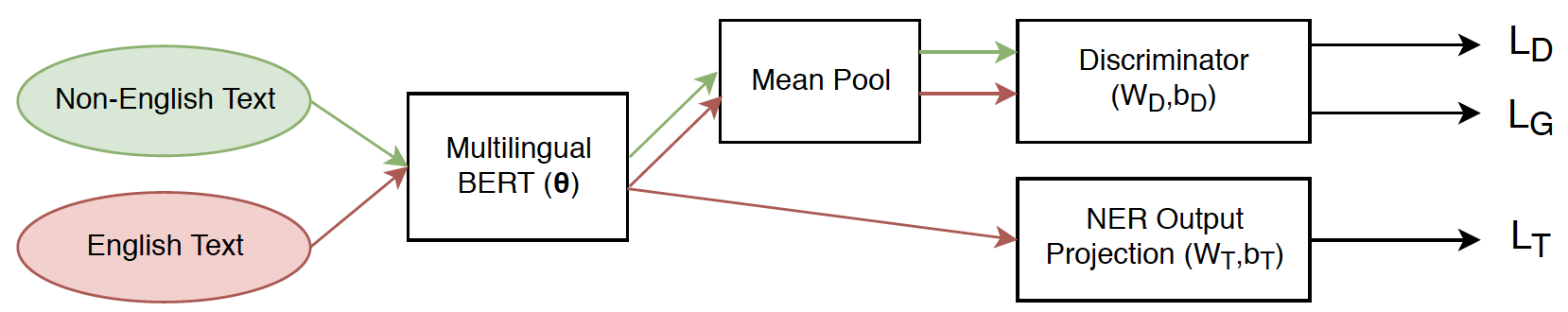}
		\caption{NER}
	\end{subfigure}
	\caption{Overview of the adversarial training process for classification and NER. All input text is in the form of a sequence of word pieces. $L_D, L_G, L_T$ refer to the discriminator, generator and task-specific losses. Parameters of each component is in round brackets.}
	\label{overview}
\end{figure*}

In this work, we explore BERT's\footnote{`BERT' hereafter refers to multilingual BERT} zero-resource performance on the multilingual MLDoc classification and CoNLL 2002/2003 NER tasks. We find that the baseline zero-resource performance of BERT exceeds the results reported in other work, even though cross-lingual resources (e.g. parallel text, dictionaries, etc.) are not used during BERT pretraining or finetuning. We apply adversarial learning to further improve upon this baseline, achieving state-of-the-art zero-resource results.

There are many recent approaches to zero-resource cross-lingual classification and NER, including adversarial learning \cite{xling_adv,adv_ner,adv_ner_2,adv_qa}, using a model pretrained on parallel text \cite{mt_encoder, interlingua, xlm} and self-training \cite{self_train_classifier}. Due to the newness of the subject matter, the definition of `zero-resource' varies somewhat from author to author. For the experiments that follow, `zero-resource' means that, during model training, we do not use labels from non-English data, nor do we use human or machine-generated parallel text. Only labeled English text and unlabeled non-English text are used during training, and hyperparameters are selected using English evaluation sets.

Our contributions are the following:

\begin{itemize}
	\item We demonstrate that the addition of a language-adversarial task during finetuning for multilingual BERT can significantly improve the zero-resource cross-lingual transfer performance. 
	\item For both MLDoc classification and CoNLL NER, we find that, even without adversarial training, the baseline multilingual BERT performance can exceed previously published results on zero-resource performance.
	\item We show that adversarial techniques encourage BERT to align the representations of English documents and their translations. We speculate that this alignment causes the observed improvement in zero-resource performance.
\end{itemize}

\section{Related Work}

\subsection{Adversarial Learning}

Language-adversarial training \cite{adv_word_trans} was proposed for generating bilingual dictionaries without parallel data. This idea was extended to zero-resource cross-lingual tasks in NER \cite{adv_ner, adv_ner_2} and text classification \cite{xling_adv}, where we would expect that language-adversarial techniques induce features that are language-independent.

\subsection{Self-training Techniques}

Self-training, where an initial model is used to generate labels on an unlabeled corpus for the purpose of domain or cross-lingual adaptation, was studied in the context of text classification \cite{self_train_classifier} and parsing \cite{self_train_parsing, self_train_parsing_xling}. A similar idea based on expectation-maximization, where the unobserved label is treated as a latent variable, has also been applied to cross-lingual text classification in \citet{em_xling}.

\subsection{Translation as Pretraining}

\citet{mt_encoder} and \citet{interlingua} use the encoders from machine translation models as a starting point for task-specific finetuning, which permits various degrees of multilingual transfer. \citet{xlm} add an additional masked translation task to the BERT pretraining process, and the authors observed an improvement in the cross-lingual setting over using the monolingual masked text task alone.

\section{Experiments}

\subsection{Model Training}

We present an overview of the adversarial training process in Figure \ref{overview}. We used the pretrained cased multilingual BERT model\footnote{https://github.com/google-research/bert/blob/master/multilingual.md} as the initialization for all of our experiments. Note that the BERT model has 768 units.

We always use the labeled English data of each corpus. We use the non-English text portion (without the labels) for the adversarial training.

We formulate the adversarial task as a binary classification problem (i.e. English versus non-English.) We add a \textit{language discriminator} module which uses the BERT embeddings to classify whether the input sentence was written in English or the non-English language. We also add a \textit{generator} loss which encourages BERT to produce embeddings that are difficult for the discriminator to classify correctly. In this way, the BERT model learns to generate embeddings that do not contain language-specific information.

The pseudocode for our procedure can be found in Algorithm \ref{pseudocode}. In the description that follows, we use a batch size of 1 for clarity.

For language-adversarial training for the classification task, we have 3 loss functions: the task-specific loss $L_T$, the generator loss $L_G$, and the discriminator loss $L_D$:
\[ L_T(y^T;x) = \sum_{i=1}^K -y^T_{i} \textrm{log}p(Y=i|x) \]
\[p(Y | x) = \textrm{Softmax}(W_T \bar{h}_\theta(x) + b_T)\]
\begin{multline*}
L_G(y^A;x) = -(1-y^A) \textrm{log}p(E=1 | x) \\
- y^A \textrm{log}p(E=0 | x)
\end{multline*}
\begin{multline*}
L_D(y^A;x) = -(1-y^A) \textrm{log}p(E=0 | x) \\
					 - y^A \textrm{log}p(E=1 | x) 
\end{multline*}
\[p(E = 1 | x) = \textrm{Sigmoid}(w_D\cdot \bar{h}_\theta(x) + b_D)\]

where $K$ is the number of classes for the task, $p(Y | x)$ (dim: $K \times 1$)  is the task-specific prediction, $p(E =1 | x)$ (dim: scalar) is the probability that the input is in English, $\bar{h}_\theta(x)$ (dim: $768 \times 1$) is the mean-pooled BERT output embedding for the input word-pieces $x$, $\theta$ is the BERT parameters, $W_T, b_T, w_D, b_D$ (dim: $K \times 768$, $K \times 1$, $768 \times 1$, scalar)  are the output projections for the task-specific loss and discriminator respectively, $y^T$ (dim: $K \times 1$) is the one-hot vector representation for the task label and $y^A$ (dim: scalar) is the binary label for the adversarial task (i.e. 1 or 0 for English or non-English).

In the case of NER, the task-specific loss has an additional summation over the length of the sequence:
\[ L_T(y^T;x) = \sum_{i=1}^K \sum_{t=1}^L -y^T_{it} \textrm{log}p(Y_t=i|x) \]
\[p(Y_t | x) = \textrm{Softmax}(W_T {h}_\theta(x)_t + b_T)\]

where $p(Y_t | x)$ (dim: $K\times 1$) is the prediction for the $t^{th}$ word, $L$ is the number of words in the sentence, $y^T$ (dim: $K\times L$) is the matrix of one-hot entity labels, and ${h}_\theta(x)_t$ (dim: $768 \times 1$) refers to the BERT embedding of the $t^{th}$ word. 

The generator and discriminator losses remain the same for NER, and we continue to use the mean-pooled BERT embedding during adversarial training.

We then take the gradients with respect to the 3 losses and the relevant parameter subsets. The parameter subsets are $\theta_D = \{w_D, b_D\}$, $\theta_T = \{\theta, W_T, b_T\}$, and $\theta_G = \{\theta\}$. We apply the gradient updates sequentially at a 1:1:1 ratio.

During BERT finetuning, the learning rates for the task loss, generator loss and discriminator loss were kept constant; we do not apply a learning rate decay.

All hyperparameters were tuned on the English dev sets only, and we use the Adam optimizer in all experiments. We report results based on the average of 4 training runs.

\begin{algorithm*}[ht]
	\caption{Pseudocode for adversarial training on the multilingual text classification task. The batch size is set at 1 for clarity. The parameter subsets are $\theta_D = \{w_D, b_D\}$, $\theta_T = \{\theta, W_T, b_T\}$, and $\theta_G = \{\theta\}$.}\label{pseudocode}
	\hspace*{\algorithmicindent} \textbf{Input:} pre-trained BERT model $h_\theta$, data iterators for English and the non-English language $L$, learning rates $\eta_D,\eta_G,\eta_T$ for each loss function, initializations for  discriminator output projection $w_D, b_D$, task-specific output projection $W_T, b_T$, and BERT parameters $\theta$
	\begin{algorithmic}[1]
		\While {not converged}
		\State $x_{En}, y^T \leftarrow \textrm{DataIterator}(En)$ \Comment{get English text and task-specific labels}
		\State $\bar{h}_{En} \leftarrow \textrm{MeanPool}(h_\theta(x_{En}))$
		\State $p^T \leftarrow \textrm{Softmax}(W_T \bar{h}_{En} + b_T)$ \Comment {compute the prediction for the task}
		\State $L_T \leftarrow -y^T \cdot \textrm{log} p^T $ \Comment{compute task-specific loss}
		\State $\theta, W_T, b_T \mathrel{{+}{=}} -\eta_T \nabla_{\theta_T} L_T$ \Comment{update model based on task-specific loss}
		\State $x_{L}, x_{En} \leftarrow \textrm{DataIterator}(L), \textrm{DataIterator}(En)$ \Comment{get non-English and English text}
		\State $\bar{h}_{L}, \bar{h}_{En}  \leftarrow \textrm{MeanPool}(h_\theta(x_{L})), \textrm{MeanPool}(h_\theta(x_{En}))$
		\State $p^D_L \leftarrow \textrm{Sigmoid}(w_D\cdot \bar{h}_L + b_D)$ \Comment {discriminator prediction on non-English text}
		\State $p^D_{En} \leftarrow \textrm{Sigmoid}(w_D\cdot \bar{h}_{En} + b_D)$ \Comment {discriminator prediction on English text}
		\State $L_D \leftarrow -\textrm{log} p^D_{En} -\textrm{log} (1-p^D_{L}) $ \Comment{compute discriminator loss}
		\State $w_D, b_D \mathrel{{+}{=}} -\eta_D \nabla_{\theta_D} L_D$ \Comment{update model based on discriminator loss}
		\State $x_{L}, x_{En} \leftarrow \textrm{DataIterator}(L), \textrm{DataIterator}(En)$
		\State $\bar{h}_{L}, \bar{h}_{En}  \leftarrow \textrm{MeanPool}(h_\theta(x_{L})), \textrm{MeanPool}(h_\theta(x_{En}))$
		\State $p^D_L \leftarrow \textrm{Sigmoid}(w_D\cdot \bar{h}_L + b_D)$ 
		\State $p^D_{En} \leftarrow \textrm{Sigmoid}(w_D\cdot \bar{h}_{En} + b_D)$ 
		\State $L_G \leftarrow -\textrm{log}(1- p^D_{En}) -\textrm{log} p^D_{L} $ \Comment{compute generator loss}
		\State $\theta \mathrel{{+}{=}} -\eta_G \nabla_{\theta_G} L_G$ \Comment{update model based on generator loss}
		\EndWhile
	\end{algorithmic}
\end{algorithm*}

\begin{table*}[th]
	\centering
	\begin{tabular}{lcccccccc}
		\toprule 
		& En & De & Es & Fr & It & Ja & Ru & Zh \\ 
		\midrule
		\citet{mldoc} & 92.2 & 81.2 & 72.5 & 72.4 & 69.4 & 67.6 & 60.8 & 74.7 \\ 
		\citet{mt_encoder} & 89.9 & 84.8 & 77.3 & 77.9 & 69.4 & 60.3 & 67.8 & 71.9 \\ 
		\midrule
		BERT En-labels & 94.2 & 79.8 & 72.1 & 73.5 & 63.7 & 72.8 & 73.7 & 76.0 \\ 
		BERT En-labels + Adv. & - & \textbf{88.1} & \textbf{80.8} & \textbf{85.7} & \textbf{72.3} & \textbf{76.8} & \textbf{77.4} & \textbf{84.7} \\ 
		\bottomrule
	\end{tabular} 
	\caption{Classification accuracy on the MLDoc test sets. We present results for BERT finetuned on labeled English data and BERT finetuned on labeled English data with language-adversarial training. Our results are averaged across 4 training runs, and hyperparameters are tuned on English dev data.}
	\label{table_mldoc}
\end{table*}

\subsection{MLDoc Classification Results}

\begin{figure}[h]
	\centering
	\begin{subfigure}[t]{0.45\textwidth}
		\centering
		\includegraphics[width=6.5cm]{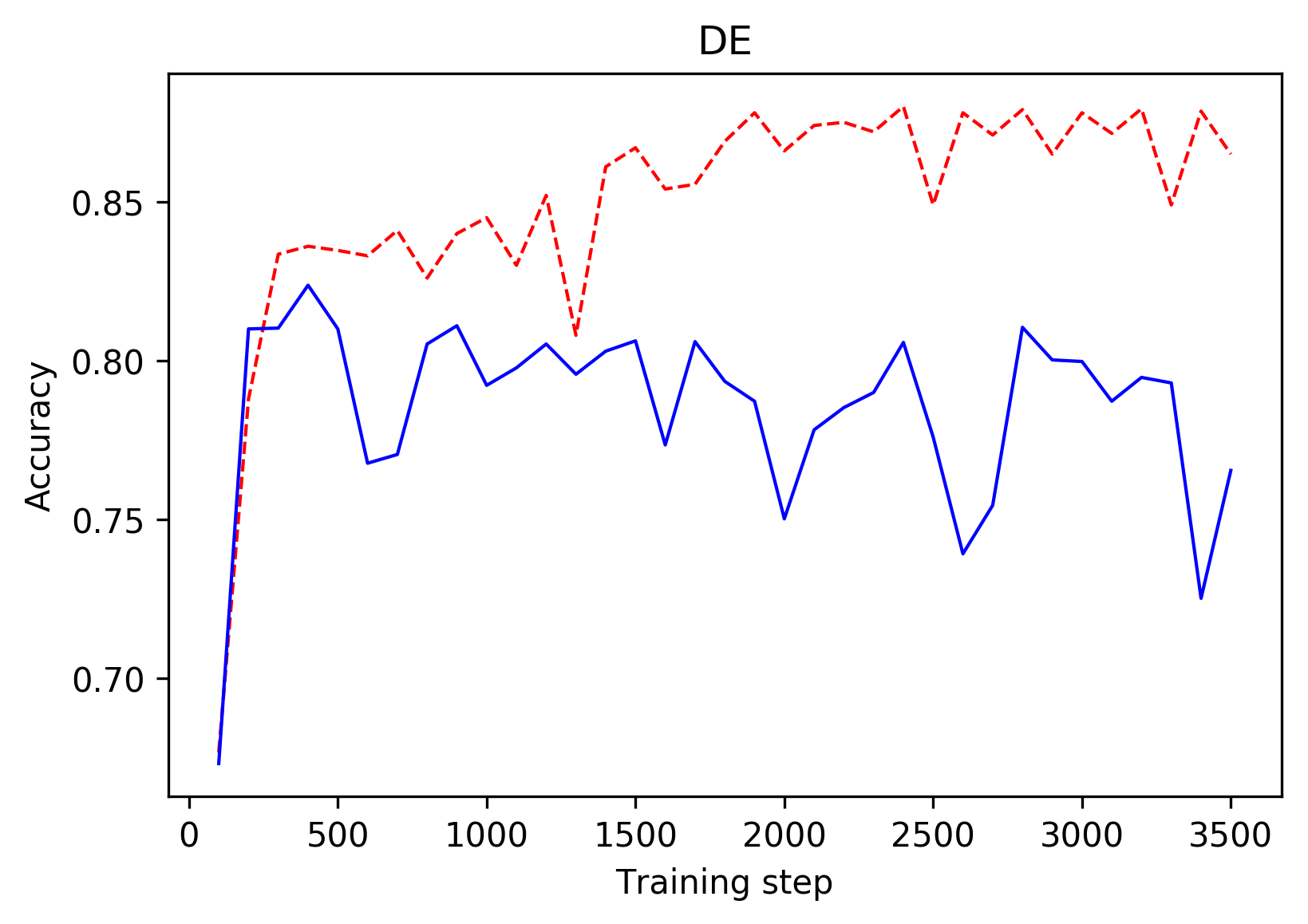}
		\caption{German test accuracy vs steps taken}
	\end{subfigure}
	\begin{subfigure}[t]{0.45\textwidth}
		\centering
		\includegraphics[width=6.5cm]{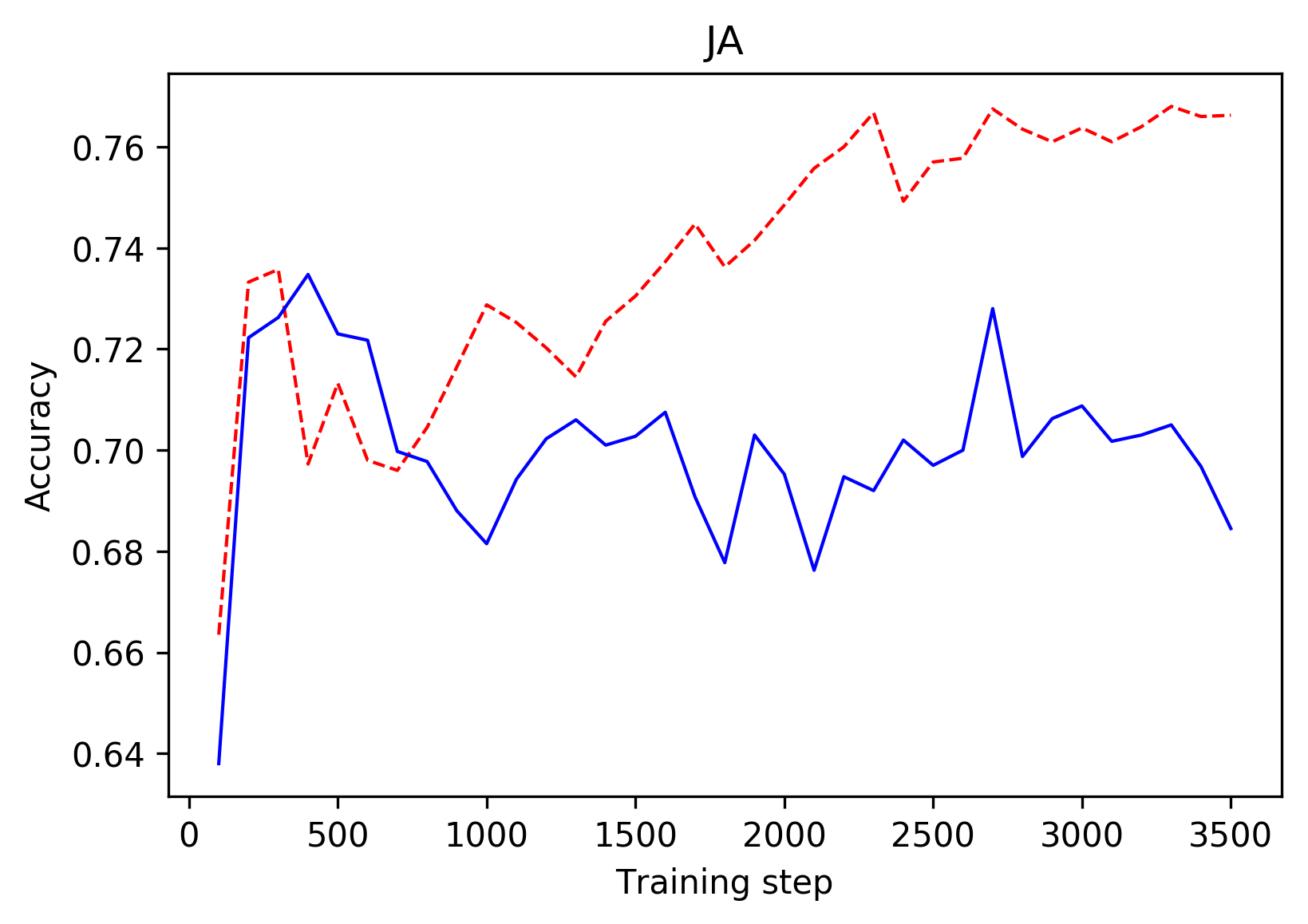}
		\caption{Japanese test accuracy vs steps taken}
	\end{subfigure}
	\caption{German and Japanese MLDoc test accuracy versus the number of training steps, with and without adversarial training. The solid line shows the performance of the non-adversarial BERT baseline. The dashed line shows the performance with adversarial training.}
	\label{fig_mldoc_progress}
\end{figure}

We finetuned BERT on the English portion of the MLDoc corpus \cite{mldoc}. The MLDoc task is a 4-class classification problem, where the data is a class-balanced subset of the Reuters News RCV1 and RCV2 datasets. We used the \verb|english.train.1000| dataset for the classification loss, which contains 1000 labeled documents. For language-adversarial training, we used the text portion of \verb|german.train.10000|, \verb|french.train.10000|, etc. without the labels.

We used a learning rate of $2\times 10^{-6}$ for the task loss, $2\times 10^{-8}$ for the generator loss and $5\times 10^{-5}$ for the discriminator loss.

In Table \ref{table_mldoc}, we report the classification accuracy for all of the languages in MLDoc. Generally, adversarial training improves the accuracy across all languages, and the improvement is sometimes dramatic versus the BERT non-adversarial baseline.

In Figure \ref{fig_mldoc_progress}, we plot the zero-resource German and Japanese test set accuracy as a function of the number of steps taken, with and without adversarial training. The plot shows that the variation in the test accuracy is reduced with adversarial training, which suggests that the cross-lingual performance is more consistent when adversarial training is applied. (We note that the batch size and learning rates are the same for all the languages in MLDoc, so the variation seen in Figure \ref{fig_mldoc_progress} are not affected by those factors.)

\subsection{CoNLL NER Results}

\begin{table*}[h]
	\centering
	\begin{tabular}{lcccc}
		\toprule 
		& En & De & Es & Nl \\ 
		\midrule
		\citet{bert} & 92.4 & - & - & - \\
		\citet{conll_mayhew} & - & 57.5 & 66.0 & 64.5 \\
		\citet{conll_ni} & - & 58.5 & 65.1 & 65.4 \\
		\citet{xling_adv} & - & 56.0 & 73.5 & 72.4 \\ 
		\citet{adv_ner_2} & - & 57.8 & 72.4 & 71.3 \\ 
		\midrule
		BERT En-labels & 91.1 & 68.6 & \textbf{75.0} & 77.5 \\ 
		BERT En-labels + Adv. & - & \textbf{71.9} & 74.3 & \textbf{77.6} \\
		\bottomrule
	\end{tabular} 
	\caption{F1 scores on the CoNLL 2002/2003 NER test sets. We present results for BERT finetuned on labeled English data and BERT finetuned on labeled English data with language-adversarial training. Our results are averaged across 4 training runs, and hyperparameters are tuned on English dev data.}
	\label{table_conll}
\end{table*}

We finetuned BERT on the English portion of the CoNLL 2002/2003 NER corpus \cite{conll}. We followed the text preprocessing in \citet{bert}.

We used a learning rate of $6\times 10^{-6}$ for the task loss, $6\times 10^{-8}$ for the generator loss and $5\times 10^{-4}$ for the discriminator loss.

In Table \ref{table_conll}, we report the F1 scores for all of the CoNLL NER languages. When combined with adversarial learning, the BERT cross-lingual F1 scores increased for German over the non-adversarial baseline, and the scores remained largely the same for Spanish and Dutch. Regardless, the BERT zero-resource performance far exceeds the results published in previous work.

\citet{conll_mayhew} and \citet{conll_ni} do use some cross-lingual resources (like bilingual dictionaries) in their experiments, but it appears that BERT with multilingual pretraining performs better, even though it does not have access to cross-lingual information.

\subsection{Alignment of Embeddings for Parallel Documents}


\begin{table}[h]
	\centering
	\begin{tabular}{cccc}
		\toprule 
		Source & Target & Without Adv. & With Adv. \\ 
		\midrule
		\multirow{7}{*}{En} & De & 0.74 & \textbf{0.94} \\ 
		& Es & 0.72 & \textbf{0.94} \\
		& Fr & 0.73 & \textbf{0.94} \\ 
		& It & 0.73 & \textbf{0.92} \\ 
		& Ja & 0.65 & \textbf{0.84} \\ 
		& Ru & 0.72 & \textbf{0.89} \\ 
		& Zh & 0.69 & \textbf{0.91} \\ 
		\bottomrule
	\end{tabular} 
	\caption{Median cosine similarity between the mean-pooled BERT embeddings of MLDoc English documents and their translations, with and without language-adversarial training. The median cosine similarity increased with adversarial training for every language pair, which suggests that the adversarial loss encourages BERT to learn language-independent representations.}
	\label{table_align}
\end{table}

If language-adversarial training encourages language-independent features, then the English documents and their translations should be close in the embedding space. To examine this hypothesis, we take the English documents from the MLDoc training corpus and translate them into German, Spanish, French, etc. using Amazon Translate.

We construct the embeddings for each document using BERT models finetuned on MLDoc. We mean-pool each document embedding to create a single vector per document. We then calculate the cosine similarity between the embeddings for the English document and its translation. In Table \ref{table_align}, we observe that the median cosine similarity increases dramatically with adversarial training, which suggests that the embeddings became more language-independent.

\section{Discussion}

For many of the languages examined, we were able to improve on BERT's zero-resource cross-lingual performance on the MLDoc classification and CoNLL NER tasks. Language-adversarial training was generally effective, though the size of the effect appears to depend on the task. We observed that adversarial training moves the embeddings of English text and their non-English translations closer together, which may explain why it improves cross-lingual performance.

Future directions include adding the language-adversarial task during BERT pre-training on the multilingual Wikipedia corpus, which may further improve zero-resource performance, and finding better stopping criteria for zero-resource cross-lingual tasks besides using the English dev set.





\section*{Acknowledgments}

We would like to thank Jiateng Xie, Julian Salazar and Faisal Ladhak for the helpful comments and discussions.

\bibliography{emnlp-ijcnlp-2019}
\bibliographystyle{acl_natbib}

\appendix

\end{document}